\documentclass{article}

\usepackage{arxiv}

\usepackage[utf8]{inputenc} % allow utf-8 input
\usepackage[T1]{fontenc}    % use 8-bit T1 fonts
\usepackage{hyperref}       % hyperlinks
\usepackage{url}            % simple URL typesetting
\usepackage{booktabs}       % professional-quality tables
\usepackage{amsfonts}       % blackboard math symbols
\usepackage{nicefrac}       % compact symbols for 1/2, etc.
\usepackage{microtype}      % microtypography
\usepackage{lipsum}		% Can be removed after putting your text content
\usepackage{graphicx}
\usepackage[numbers]{natbib}
\usepackage{doi}
\usepackage{alphabeta}

\usepackage{authblk}
\usepackage{tipa}
\usepackage{amsmath}
\usepackage{xcolor}

\title{LIMBA: An Open-Source Framework for the Preservation and Valorization of Low-Resource Languages using Generative Models}

\author[1, 2, 5]{Salvatore Mario Carta}
\author[2]{Stefano Chessa}
\author[3]{Giulia Contu}
\author[2]{Andrea Corriga}
\author[2]{Andrea Deidda}
\author[1]{Gianni Fenu}
\author[3]{Luca Frigau}
\author[1]{Alessandro Giuliani}
\author[5]{Luca Grassi}
\author[1]{Marco Manolo Manca}
\author[1]{Mirko Marras}
\author[3]{Francesco Mola}
\author[6]{Bastianino Mossa}
\author[4]{Piergiorgio Mura}
\author[3]{Marco Ortu}
\author[1]{Leonardo Piano}
\author[4]{Simone Pisano}
\author[1]{Alessia Pisu}
\author[1]{Alessandro Sebastian Podda}
\author[1]{Livio Pompianu}
\author[2]{Simone Seu}
\author[1]{Sandro Gabriele Tiddia}

\affil[1]{Department of Mathematics and Computer Science, University of Cagliari}
\affil[2]{VisioScientiae S.r.l., Cagliari}
\affil[3]{Department of Economics and Business Sciences, University of Cagliari}
\affil[4]{Department of Humanistic Studies, University for Foreigners of Siena}
\affil[5]{The Cloud Alchemist S.r.l., Cagliari}
\affil[6]{FASI: Federation of Sardinian Associations in Italy}
\date{}

\begin{document}
\maketitle

\begin{abstract}
Minority languages are vital to preserving cultural heritage, yet they face growing risks of extinction due to limited digital resources and the dominance of artificial intelligence models trained on high-resource languages. This white paper proposes a framework to generate linguistic tools for low-resource languages, focusing on data creation to support the development of language models that can aid in preservation efforts. Sardinian, an endangered language, serves as the case study to demonstrate the framework's effectiveness. By addressing the data scarcity that hinders intelligent applications for such languages, we contribute to promoting linguistic diversity and support ongoing efforts in language standardization and revitalization through modern technologies.
\end{abstract}


\section{Introduction}
\label{sec:intro}
Languages, including those spoken by minority groups, represent a priceless cultural heritage that reflects the history, traditions, and identity of a people. 
The preservation and enhancement of minority languages are of paramount importance for maintaining linguistic diversity and safeguarding the distinctive knowledge associated with the communities that speak them. Each language represents a unique way of interpreting the world, and the loss of a language results in the loss of this perspective. 
Promoting minority language rights and intercultural dialogue contributes to social cohesion and the richness of the global human heritage.

The increasing importance of generative models, including those based on artificial intelligence, is evident in numerous domains. These models are utilized in various applications, including the generation of text, images, and music, as well as in more practical contexts such as machine translation, code completion, and virtual assistance. However, these models are primarily developed and optimized for high-resource languages, such as English, Chinese, French, and numerous other European languages, which have extensive training data. 
This creates a disparity between minority and less documented languages, which frequently remain less supported and benefit less from AI technologies. Overcoming this challenge could enhance equitable access to technology and promote greater language inclusivity globally.

The aforementioned models necessitate the availability of substantial quantities of data to ensure effective training, a requirement that presents a considerable challenge for endangered languages. 
These languages are afflicted by a severe scarcity and fragmentation of available data, with few written or digital resources and a steadily decreasing community of speakers. The paucity of data makes it challenging to develop high-quality language models for these languages, thereby exacerbating the risk of cultural and linguistic extinction. 
It is imperative to address this data deficit in order to preserve and enhance endangered languages, which can be achieved by promoting initiatives to collect and digitize texts, recordings, and other language resources.

Indeed, in recent years, research has been aimed at narrowing the gap between more widely spoken languages and those that are endangered by seeking methods of revitalization and enhancement \cite{olko2021revitalizing}. 
In the context of information technology, where these languages are frequently designated as low-resource languages due to the paucity of available data, considerable emphasis has been placed on the creation of high-impact artificial intelligence-based linguistic resources that could facilitate the preservation and dissemination of lesser-known languages \cite{mohanty2024applying}. 
The generation of the described tools is, therefore, constrained by the availability of data rather than by the complexity of development. Consequently, several entities, including Google and Mozilla, have initiated programs to encourage data collection. \textit{Google's Project Euphonia} and \textit{Mozilla Common Voice}, for instance, encourage the crowdsourcing of voice data, whereby native speakers can contribute by recording their voice.

This paper proposes a methodology, still under development, for generating a set of linguistic tools capable of constructing new data in little-used languages so that they can be used to train a language model. In order to evaluate the effectiveness of the work presented, the model generation pipeline is tested on Sardinian, a language that, due to its complexity and risk of disappearance \cite{mensching2000internet}, is a perfect case study for the set task. 
In fact, this language is often the cause of cultural and political debates regarding its standardization, an issue that has aroused the curiosity of several researchers \cite{mereu2021efforts}. The application of a methodology such as the one described in this paper could even facilitate the task of standardizing the language and defining its sociolinguistic situation more clearly. 

The main objectives of this work are therefore: 

\begin{itemize}
    \item defining an end-to-end framework for low-resource languages to generate a language model by defining a set of tools suitable for analyzing and generating data in a minority language; 
    \item testing the effectiveness of the framework using the Sardinian language as a case study.
\end{itemize}

The remainder of the paper is organized as follows: Section \ref{sec:background} will give an overview of existing work in this context, Section \ref{sec:methodology} will describe the proposed work in detail, Section \ref{sec:considerations} will outline some considerations regarding the impact of the devised framework, and finally Section \ref{sec:conclusions} will conclude the paper.

\section{Background}
\label{sec:background}
This section, aimed at providing an adequate context, describes the state of the art of low-resource languages and their development in the field of language processing. An overview of the existing literature in linguistics on these languages will be given, and then it will be described how they have been treated from a computer science perspective. In addition, it will be highlighted how the Sardinian language has been related in these contexts.  

\subsection{Foundations of Low-Resource Languages}
%The following provides background on the definition of a low-resource language, the characteristics of such a language, and the steps that have been taken to preserve and improve it. 
In the following, we provide a proper background, from giving a proper definition and the main characteristics of low-resource languages to describing the actions taken for their preservation and improvement.

\subsubsection{Preliminaries}
A \textit{low-resource language} refers to a language for which there is a lack of digital data, such as text or audio recordings, useful for the development of advanced linguistic technologies, e.g., speech recognition or machine translation. Therefore, the development of tools and applications that can effectively incorporate it is limited. 
In summary, such languages can be interpreted as those with the least information and low density~\citep{cieri2016selection}. It is clear that such languages present numerous challenges, including the difficulty of developing accurate Artificial Intelligence (AI) models, the lack of annotated corpora, and the limited technological support for digitization. In addition, the scarcity of language documentation, and often the lack of financial or institutional resources to preserve it, makes it even more difficult to make the language more accessible and usable in modern technologies. 
%\citet{magueresse2020low} provide an overview of what has been done to improve the situation of these languages and what future challenges lie ahead. 
\citet{magueresse2020low} provide an overview of the recent improvements and the future challenges.

\subsubsection{Languages Taxonomy}
Linguistic taxonomies symbolically represent the hierarchical relationships between terms or entities in a language. Although they are widely used in a variety of contexts, manually updating and maintaining them is a complex process and difficult to scale. As this procedure is even more complicated for minority or low-resource languages, a process of taxonomy enrichment is often undertaken. In the work of \citet{takeoka2021low}, pre-trained language models are used to compensate for the lack of information in low-resource environments.

From this point of view, the \textit{Sardinian} language has an extremely complicated taxonomy, characterized by several variants that are still being defined and by great dialectal diversity. 
Sardinian is a Romance language that emerged during the Romanization of Sardinia in the 3rd century BC. Several experts have classified Sardinian as an Italo-Romance language due to its affinity with Italian as an administrative and cultural reference~\cite{pellegrini1977carta}. 
Nevertheless, due to its unique linguistic characteristics and structural distance from other Romance languages~\cite{merlo2009italia,wagner1921ländliche,marzo17}, Sardinian, nowadays, is considered a distinct linguistic branch, as it also shares traits from both Eastern and Western Romània, impeding a precise categorization within the Romance family~\cite{virdis03a}. These characteristics make this language an ideal study case for work in description. 

\subsubsection{Preservation and Valorization}
Preserving and promoting lesser-used languages is critical to safeguarding the world's cultural and linguistic diversity. These languages often convey the traditions, histories, and identities of minority communities, which are at risk of disappearing without appropriate intervention. Valuing these languages also promotes social inclusion and ensures that even the smallest communities have access to modern technological tools. Investing in their documentation and digitization is essential to prevent the loss of unique knowledge, thereby contributing to a richer and more diverse global heritage.
Many scholars in the field have worked to find new ways to keep such languages alive and prevent them from disappearing. \citet{bird2020decolonising}, for example, suggest new ways of working with Indigenous communities; similarly, \citet{muzoora2014valorisation} highlight the challenges of language policy in the context of African education, documenting how the educational aspect occupies a prominent position in the process of strengthening and preserving a language.

Even for the Sardinian language, it seems unquestionable to state an ongoing process of revalorization of the language in the perception of the island's inhabitants~\cite{valdes2007capitolo,mura2024attitudes} and the diffusion of Sardinian into new domains, often public and formal~\cite{marra2012lingue,ghimenton2016ideologies,russo2016sardinian, linzmeier2021visibilita,masullo2021dialetti}, apparently - although this point still needs to be further investigated by specialized studies - with some linguistic features of common diffusion and partially distant from those observable in diaphasically \textit{low} Sardinian. What is taking place can perhaps be classified, at least if we take into consideration certain areas of Sardinia, in particular the major urban centers, as a process of diacrolectia, i.e., a situation in which the local language is increasing its use in \textit{high} contexts, but remains in great difficulty in the spheres of everyday conversation, now the preserve of the national language, a fact that still places Sardinian in a serious condition of danger of extinction.

\subsection{Low-Resource Language Processing}
The following sections describe the background of Natural Language Processing (NLP) in the context of data-poor languages, with the aim of presenting state-the-art in terms of data collection, tools to support language analysis, processing of audio material in minority languages, translators, and finally, language models.  

\subsubsection{Resource Collection}
As mentioned above, one of the main challenges for those doing research in low-resource languages is the scarcity of digitized language resources, such as corpora, audio recordings, or annotated texts. Many of these languages are spoken by small communities or in remote areas, making it difficult to collect data systematically. In addition, a lack of standardization and language tools, such as dictionaries or natural language processing models, further complicates the process. It is therefore crucial to develop innovative data collection strategies that use adaptive technologies and actively involve local communities to ensure that languages are adequately represented in the digital landscape.

The proposal of \citet{madaan2020practical} describes a method for collecting high-quality data for low-resource languages to be used to train a machine translation system from a more widely used language (English) to a minority language (Hindi). The method used is based on obtaining captions for images that illustrate concepts relevant to most of the world's languages. In other cases, as in the proposal of \citet{de2024data}, a custom data collection pipeline is used to automate and streamline the process of building text corpora from the Web for low-resource languages. Of course, text corpora can also be extracted from dialogues between people; in this context, the work of \citet{yusupujiang2021data} proposes a method for generating a dialogue corpus for Uyghur, but generalizable to other low-resource languages. 

As with other minority languages, several data collection projects have focused on Sardinian; \cite{pisano2022appunti} describe a corpus of contemporary Sardinian collected from interviews with Sardinians living outside Sardinia. However, the annotation and dataset creation process is done manually, leaving room for errors and requiring a lot of resources and time. Therefore, there is an urgent need to develop algorithms that perform this process automatically based on reliable and powerful AI systems. 

\subsubsection{Linguistic Modeling}
The Linguistic Modeling task considered in this paper refers primarily to the grammatical analysis tools used to study linguistic details for languages that are still under study. Such tools automatically annotate portions of text, facilitating the work of a linguist; their use can also aid teaching and assessment tasks for uncommon language knowledge. This paper focuses on three main tools that the Sardinian language, among other languages, urgently needs: a Part-of-Speech tagger, a lemmatizer, and a language variant identifier. 

\paragraph{\textit{Part-of-Speech Tagger}}
A Part-of-Speech (PoS) tagger is an NLP tool that assigns a grammatical category to each word in a sentence, such as a noun, verb, or adjective. It helps to identify the function of each word in the context of the sentence, making syntactic analyses easier. Such a tool can be critical in developing low-resource languages, as it automates linguistic analysis even in contexts lacking structured linguistic resources. With the PoS tagger, the creation of annotated corpora and translation tools can be accelerated, contributing to the preservation and development of these underrepresented languages.

Several Machine Learning algorithms have been used to develop these kinds of tools. \citet{mercer1992class} and \citet{christodoulopoulos2010two} exploit clustering models to annotate each word and assign a label. Another technique often used involves Hidden Markov Models (HMMs) approaching the tagging problem as a sequence-to-sequence problem; for example, some models are trained on a “native” language whose annotated data are available, then applying a grounding step to obtain the annotations in the language of interest \cite{buys2016cross,cardenas2019grounded}. Finally, some works interpret PoS tagging as a classification problem; among them, \citet{duong2014can} proposed a softmax classifier trained on projected English-language annotations and subsequently adjusted to the tags of the language of interest. However, the performance of such a classifier was enhanced in further works by using bidirectional Long Short-Term Memory models~\cite{fang2016learning,fang2017model}.

In the context of the Sardinian language, to the best of our knowledge, there is no automatic annotation tool like those described above. An automated PoS tagger prototype is currently under development~\cite{mura2024}. In particular, the model is generated by fine-tuning a small language model (BERT) adapted for the token classification task on a manually annotated Sardinian language dataset.

\paragraph{\textit{Lemmatizer}.}
A \textit{lemmatizer} is an NLP tool that reduces words to their basic form or lemma, eliminating inflection and grammatical variants. This process helps to unify terms in the text, thereby improving the accuracy of linguistic analysis. Such a tool can be crucial for developing low-resource languages, as it simplifies linguistic standardization even in languages with complex variants. The lemmatizer can improve the quality of tools such as dictionaries or translation models, thus accelerating the creation of language resources.

Some research has focused on training specific models to perform the task of lemmatization. \citet{yarowsky-etal-2001-inducing} proposed, for instance, a training process for a multilingual lemmatizer. This procedure comprises a bootstrap phase of the system, initiated from the projections of one language onto a second language of interest. The strength lies in its capacity to process noisy and incomplete initial projections.
The generation of lemmatizers also employs editing-based strategies that utilize Long-Short-Term-Memory (LSTM) models, as they are optimally suited to handle text sequences~\cite{makarov2018neural}.

The absence of an automatic lemmatizer in this language confirms the adherence of the case study in Sardinian. However, an ongoing project~\cite{pisano2022appunti} aims to manually annotate and lemmatize dialogue parts to create a training dataset for a Sardinian-language lemmatizer. 

\paragraph{\textit{Language Variant Identifier}.}
A \textit{language variant identifier} is a tool designed to detect and distinguish language variants, such as dialects or regional forms, within a text. This enables the precise analysis of multivariant texts, adapting language resources to the specific context. Such a tool is invaluable for the languages described in this paper, as it enables the mapping and preservation of local, poorly documented variants. Recognizing and classifying these variants can facilitate the development of richer and more representative language resources.

Such tools are not particularly widespread as not all languages possess profoundly different variants from one another. However, some works have attempted to provide a direction for identifying minority languages~\cite{king2015practical}. Other works have proposed algorithms based on Markov models~\cite{dunning1994statistical} or Monte Carlo methods~\cite{poutsma2002applying} for language identification in multilingual texts. The use of high-performance technologies for this task would facilitate the massive acquisition of data from sources such as the Web or large databases, a task not suitable for manual work. 

The case of the Sardinian language is unique; this language has two macro-variants, Logudorese and Campidanese, but several dialects are classified as \textit{middle languages} (Mesanía). This fragmentation of the language makes all language-processing tasks difficult. However, an identifier could help distinguish cases and increase the performance of the automatic models described so far and those described later. 

\subsubsection{Speech Processing}
Speech Processing is a field of NLP that deals with analyzing, interpreting, and generating spoken language. It includes two main technologies: \textit{speech-to-text }(s-t-t), which transcribes speech into written form, and \textit{text-to-speech} (t-t-s), which transforms a written text into a synthetic voice. These tools make it possible to create speech interfaces, improve accessibility, and automate speech understanding.
This area of language processing represents a strategic resource for minority languages, as the creation of s-t-t and t-t-s models, even for languages with limited digital resources, can contribute to their preservation and dissemination. It also improves access to educational and communicative tools for communities that use these languages, increasing their presence in the digital world.

One of the most popular approaches to construct a speech-to-text for a resource-poor language is integrating a translator into the audio-to-text transformation process. This method was adopted in the work of \citet{bansal2018low}, in which a neural encode-decoder model learns to translate foreign speech directly in a scenario where data and computational resources are scarce. Similarly, it has been shown that automatic speech-to-text translation improves significantly for data-poor languages by pre-training a model on high-resource language data \cite{stoian2020analyzing}. 
The t-t-s procedure, which represents a reversal of the aforementioned methodology, is inherently more complex. Indeed, the objective is to generate audio from written text. End-to-end models for low-resource languages are discussed in the literature. The work of \citet{tu2019end} demonstrates that data scarcity can be overcome by constructing t-t-s models using transfer learning techniques, whereby knowledge is transferred from a high-resource language. Research in this area also provides guidance and advice for building speech synthesis models for data-poor languages \cite{kazantsevaaspeech}. 

The Sardinian language is devoid of any tools for audio processing. It is, therefore, imperative that performant resources be generated with the utmost urgency to prevent the technological gap between Sardinian and other languages from widening further.

\subsubsection{Machine Translation}
\textit{Machine translation} is a technology that employs algorithms to automatically translate text from a source language to a target language, thereby facilitating communication and access to information between different languages. This process is based on language models that analyze the structure and meaning of text from a source language and produce a translation into a target language, striving to maintain consistency and accuracy.
The translation of text from low-resource languages, which are often less well documented and have few digital resources, facilitates the creation of cultural and linguistic bridges, thereby enabling their integration into the global digital landscape. Furthermore, machine translation not only contributes to the preservation and development of these languages by making content available in multiple languages and promoting greater linguistic inclusiveness but is also an indispensable tool for generating new data simply by using texts from languages other than the target language. 

As in the case of speech processing, transfer learning is often used as a training method for translation models. In the proposal of \citet{zoph2016transfer}, this strategy is applied to a neural machine translation model to improve BLEU scores (a common evaluation metric in translation tasks) in several low-resource languages. However, other approaches propose the introduction of a prior model to training; \citet{kumar2019augmented}, for example, adopted the Zero-Shot Translation technique to train an encoder model, whereas, in the work of \citet{baziotis2020language}, a Language Model adds a regularization term allowing the predictions of a neural machine translation model to be validated. This strategy can be interpreted as a distillation technique; in essence, the language model teaches the neural model the target language. 

\citet{tyers2017rule} describe the process of generating an automatic translation system from Sardinian to Italian. The authors exploit a Rule-Based Machine Translation technology to train the translator and use metrics such as Word Error Rate (WER) to evaluate its performance. In a follow-up study, the same authors provided an overview of the improvements and extensions to other languages of the created translator~\cite{khanna2021recent}. Their work also illustrates how a machine translation platform for low-resource languages can be crucial to language technology access, preventing minority languages from being ignored.

\subsubsection{Generative Modeling}
Generative models are Artificial Intelligence systems that can create new data from learned patterns, such as text, images, or sounds. These models learn from existing information and generate original content that follows the same rules or styles as the source material, making them useful for many creative and analytical applications. In this paper, we will focus primarily on generative language models, which can be a revolutionary resource in the context of low-resource languages. These can produce texts and resources in low-resource languages, expand the available corpus, and support the creation of educational, cultural, and technological content. They also enable the development of advanced linguistic tools for these languages, contributing to their preservation and dissemination in digital and non-digital environments. 

The availability of a large number of data strongly constrains the construction of models of this type. As pointed out earlier, this fact severely limits the training of generative models for low-resource languages. However, some studies have proposed methods of constructing datasets suitable for pre-training Large Language Models (LLMs) in minority languages, e.g., a dataset containing high-quality material for the Hindi language~\cite{parida2024building}. Similarly, for the Sámi language, in the work of \citet{paul2024towards}, Web resources are provided in order to create a clean dataset for training language models, even comparing some LLMs on the interpretation ability of this language. 
In some cases, one approach to achieving generative artificial intelligence tools in low-resource languages is to “teach” large multilingual language models new languages using In-Context Learning \cite{cahyawijaya2024llms} or Fine-tuning \cite{lankford2023adaptmllm} techniques. 

However, building an accurate language model for low-resource languages is still an open field of research and is under study. To the best of our knowledge, no end-to-end pipelines describe step-by-step how to build new data and train language models. 
The Sardinian language, like the remaining minority languages, also suffers from the lack of a dedicated language model. Although some LLMs, such as GPT4 and Llama3, are able to generate some parts of the Sardinian text, such text is often grammatically incorrect, inaccurate, and, in some cases, even made up. It is expected that a model utterly dedicated to the Sardinian language, although smaller in size than the models mentioned before, can achieve significantly higher performance.

\section{Methodology}
\label{sec:methodology}
This section describes the devised framework, providing a general overview by presenting the high-level architecture and, subsequently, the details of each component.

\subsection{Modules and Communication}
\begin{figure*}[h!]
    \centering
    \includegraphics[width=1\textwidth]{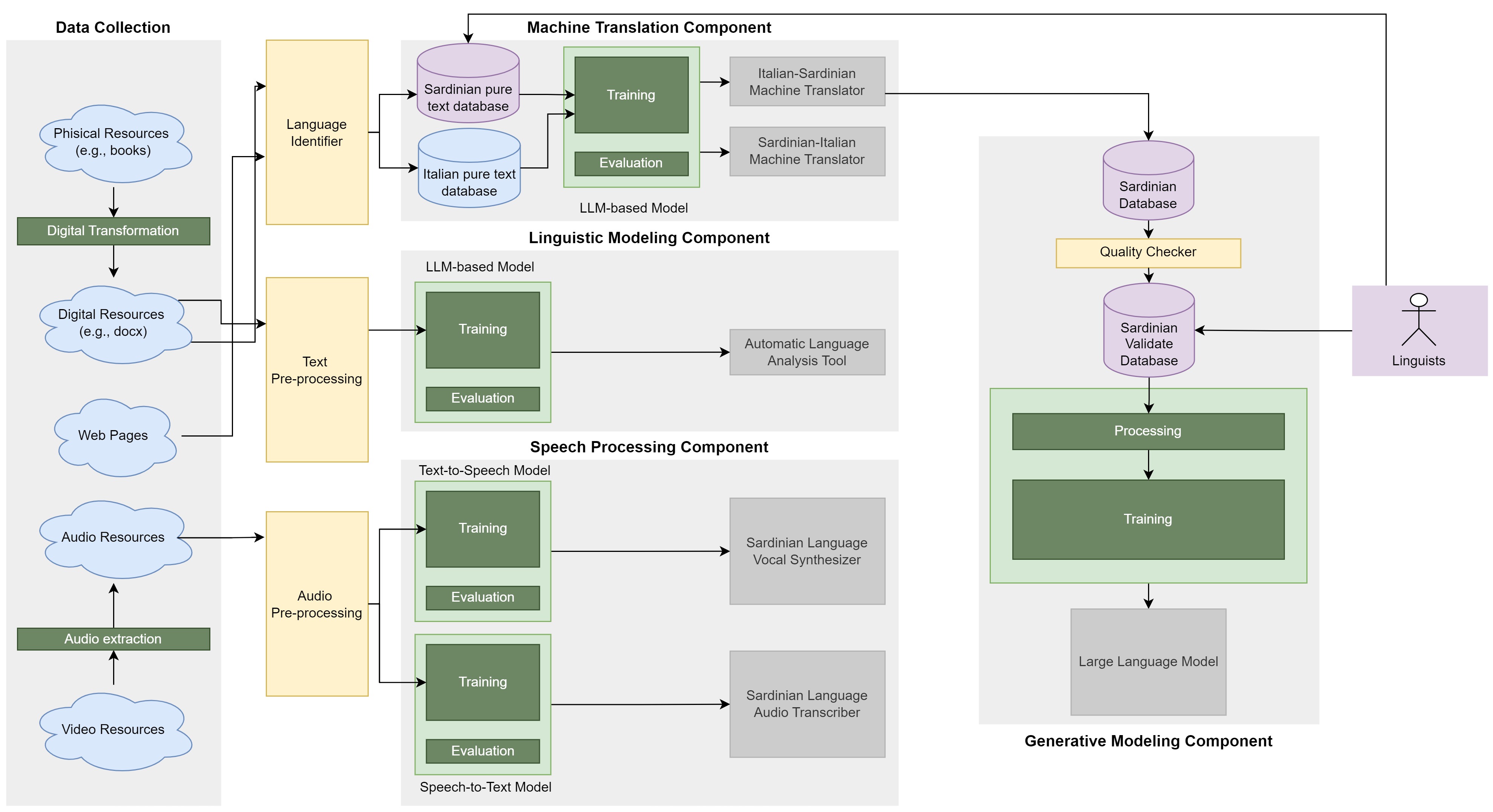}
    \caption{\textbf{Framework Architecture}}
    \label{fig:pipeline}
\end{figure*}

Figure~\ref{fig:pipeline} depicts the architecture of the developed framework, which is characterized by the concatenation of multiple modules, each producing a result in its own right. Hence, in addition to providing a method for generating a language model for low-resource languages, AI-based tools suitable for developing and enhancing the language of interest are described and developed.

The system consists mainly of 5 blocks: 

\begin{itemize}
    \item \textbf{Data collection}: This block collects written text and audio data from heterogeneous material, processes it, and makes it available for training AI models;
    \item \textbf{Linguistic Modeling}: Grammatical analysis models are developed for the target language at this stage. Let us point out that, in this paper, we limit our discussion to the PoS tagging task;
    \item \textbf{Machine Translation}: The block is aimed at training and evaluating translators from a minority language to a high-resource language;
    \item \textbf{Speech Processing}: This pipeline component is designed to develop intelligent models capable of performing \textit{speech-to-text}, i.e., creating text contents from input audio, and text-to-speech, i.e., generating audio by receiving written text as input;
    \item \textbf{Generative Modeling}: Once the preceding blocks have been consolidated, they are used to acquire sufficient data for training a language model in the target minority language. 
\end{itemize}

Furthermore, to ensure that the input data are compatible with the intended task, the pipeline incorporates two cross-modules, i.e., the \textbf{Language and Variant Identifier} and the \textbf{Language Quality Checker}. The former filters the received data, labeling it among the various potential variants or other prevalent languages it intends to interact with, whereas the latter is aimed at discarding data of poor quality to lessen the effort of domain experts (linguists or glottologists) who assist in evaluating the quality of the system's output and input. 

In the case of the Sardinian language, these two tools are essential. As previously stated, Sardinian is a highly fragmented language that necessitates filtering, labeling according to a macro-variant, and final assessment of the quality and consistency of its texts and audio. 

\subsection{Cross-Module Components}

This part outlines how the cross-module components mentioned above will be created. 

\subsubsection{Language and Variant Identifier}
\label{sec:variant_identifier}
In addition to facilitating data filtering, as previously discussed, the language identification tool enables the categorizing of low-resource languages within their respective variants, allowing domain experts to access more detailed language analyses and identify aspects of the language that are not immediately apparent. From a computer science perspective, developing a language identifier involves finding a tool capable of assigning a label to a block of text (it should be noted that, in this paper, we are limited to language identification from text)%. We are attempting to solve what is referred to as a text classification problem. 
, i.e., the problem is addressed in a \textit{classification} scenario.
The proposal we present to address this problem is to construct a classifier based on pre-trained language models.

Given a dataset $\mathcal{T} = \{(t_j, l_j^k) | \ j=1,\dots,n ; \ l_j^k \in \{l^1, \dots, l^m\} \}$, where $t_j$ represents a chunk of text, $l_j^k$ represents the language, chosen from a finite number of languages, in which the text is written, the goal is to train a classifier that takes as input a generic text $t$ and returns its correct language label. One potential methodology for achieving this objective is fine-tuning a pre-trained language model. This entails utilizing the dataset $\mathcal{T}$, or a subset thereof, to instruct the selected model in the target language. Considering the availability of computing resources, a selection of larger or smaller models may be made. 
In the case of the Sardinian language, experiments are currently underway to assess the efficacy of the BERT \cite{devlin2018bert} language model in classifying the three macro-variants that comprise the Sardinian language. The selection of this model was guided by its relative "lightness" in comparison to other LLMs, which allows for a more streamlined and comprehensive fine-tuning phase. However, it is expected to extend the number of labels for this model, i.e., to teach the model to distinguish Sardinian from other similar languages, such as Italian and Spanish, so these types of languages can be used for translation tasks. 

\subsubsection{Language Quality Checker}
The quality checker module is indispensable for building high-performance, high-quality models. When used during the training phase of language models, it has been shown that high-quality data can increase their accuracy \cite{xie2023data}. 

Accordingly, we introduce a proper strategy to determine whether data meets quality standards. Such a task can be interpreted mathematically as a binary classification problem, i.e., given a set $\mathcal{T}=\{t_j | j=1,\dots,n\}$, where $t_j$ represents a text chunk, we want to estimate a function $f$ such that: 

\begin{equation}
    f(t) = 
    \begin{cases}
        \texttt{high-quality} & \text{if $t$ is an high-quality text chunk} \\
        \texttt{low-quality} & \text{if $t$ is a low-quality text chunk}
    \end{cases}
    \hspace{.5cm} \text{for } t \in \mathcal{T}.
\end{equation}

Similar to the Language and Variant Identifier described in Section~\ref{sec:variant_identifier}, we propose to fine-tune a BERT model, suitably adapted to binary classification, to classify text. To do this requires a domain expert to manually annotate each text chunk and thereby create a dataset of the type $\mathcal{T}= \{(t_j, q_j) | j=1,\dots,n ; q_j \in \{\texttt{high-quality}, \texttt{low-quality}\}\}$, where $t_j$ represents a text chunk and $q_j$ the associated qualitative degree. 
The latter would then represent the training set of the proposed model. 
With the correct data amount and training setting, this model can optimally discriminate input data by filtering them by quality.

\subsection{Linguistic Modeling Component}
This block represents the initial phase of generating a language model for a low-resource language. In detail, a portion of the collected data will be analyzed grammatically, providing insights into the language structure and offering practical assistance to scholars engaged in this field of study. Let us remark that, although in this paper we will focus on the PoS tagging problem, other grammatical analysis tools can be constructed following the guidelines outlined below. 

\subsubsection{Problem Formalization}
Mathematically, the aim is to define a function capable of assigning a label to each word in a sentence that describes the grammatical category of the word. Given a sentence $\mathbf{s}$, representable as a vector of the type $[s_1, \dots, s_m]$ in which each entry $s_k$ is a word or punctuation of the sentence $\mathbf{s}$, and a list of PoS tags $\mathcal{U}$, the goal is to estimate a function $f$ such that:

\begin{equation}
    f(\mathbf{s}) =
    \begin{cases}
        (s_1, u_1) \\
        \dots\\
        (s_m, u_m)
    \end{cases}
    \hspace{.5cm} \text{where } u_1, \dots, u_m \in \mathcal{U}.
\end{equation}

This function associates each word with the correct tag, selected from the list $\mathcal{U}$.

\subsubsection{Resource Collection}
In order to develop the model, data collection must be conducted by a domain expert, such as a linguist or glottologist. This individual is responsible for identifying which sentences are most effective at teaching an AI model the grammatical features of a language. Additionally, the data set must include consistent tags for each class. This approach helps to ensure that the training dataset is not imbalanced towards specific tags, which could negatively impact model performance and lead to overfitting issues. 

\subsubsection{Resource Pre-Processing}
Subsequently, the collected resources must be transformed in order to be fed to the artificial intelligence algorithms. In particular, it is necessary for a domain expert to manually create a training dataset that contains an appropriate number of sentences. Each sentence must be broken into words or punctuation marks, and each must be associated with the correct tag. In this way, the dataset can be used to teach a model to correctly tag parts of text. 

Our contribution, concerning the Sardinian language, is the tagging of approximately 700 sentences by a domain expert, which were subsequently imported and stored into a proper digital structure through the \textit{datasets}\footnote{\url{https://pypi.org/project/datasets/}} Python package.
In doing so, we selected the \textit{Universal POS Tag}\footnote{\url{https://universaldependencies.org/u/pos/}} as the tag set, and each word in the sentence was tokenized through the BERT model tokenizer to be properly prepared for the next step. Finally, the collected data set was divided into two parts: one set to be used for training the model and one to evaluate its performance. 

\subsubsection{Model Development}
The subsequent phase is model training. We propose to utilize a multi-class classification model that is, however, capable of interpreting the written text and extracting information from the relationships of the words that comprise a sentence. Once more, the BERT model, renowned for its bidirectional capabilities, represents an optimal choice for fulfilling the tagging task. Furthermore, the number of weights that comprise the model is relatively modest in comparison to other models, which enables effective fine-tuning even in the absence of substantial computational resources.

It is worth pointing out that in our experiments related to the Sardinian language, we have fine-tuned various BERT models, including the BERT model in its \textit{large} version, using the dataset described in the previous section. The configuration employed is outlined in Table~\ref{tab:tuning_setting}. 

\begin{table}[h!] 
	\centering
	\caption{Setting used for fine-tuning the BERT model for the PoS tagging task. }
    \label{tab:tuning_setting}
    \begin{tabular}{lc}
    \hline
        \textbf{Hyper-Parameter}& \textbf{Value}\\
    \hline
        Epochs & 50\\
        Batch size&16\\
        Learning Rate&2e-5\\
        Weight decay&0.01\\
        Training Loss& Cross-Entropy Loss\\
    \hline
    \end{tabular}
\end{table}

\subsubsection{Model Evaluation}
In order to evaluate a model in the PoS tagging task, it is essential to employ appropriate metrics. Given that the problem has been defined as a multi-class classification task, relying on the metrics defined for such problems, namely Recall, Precision, and F1-score \cite{grandini2020metrics}, is appropriate.
For the experiments described above, the aforementioned metrics were calculated on a set of approximately 200 sentences, yielding excellent values of approximately 89\% for all metrics. Nevertheless, comprehensive methodologies to enhance the values of these metrics are still under investigation. 

\subsection{Machine Translation Component}
The next step is to create a model capable of producing a large number of text data to create a database sufficient for generating a language model. For low-resource languages, a clever way to procure such data is to take advantage of a translator that transforms texts in a high-resource language into the target language. In this section, we aim to explain how to build an effective machine translator; such methodology also covers the scenario where the goal is to translate text from a low-resource language to a high-resource language. 

\subsubsection{Problem Formalization}
From a formal point of view, a translator from an $\mathcal{L}_1$ language to an $\mathcal{L}_2$ language can be interpreted as follows. Given a set of texts $\mathcal{T}_1$ in the $\mathcal{L}_1$ language and a set of texts $\mathcal{T}_2$ in the $\mathcal{L}_2$ language, with $\mathcal{L}_1 \neq \mathcal{L}_2$, a translator is a function f such that: 

\begin{equation}
\begin{aligned}
    f: & \, \mathcal{T}_1 \to \mathcal{T}_2 \\
       & \, \mathbf{s^1} \mapsto \mathbf{s^2}
\end{aligned}
\end{equation}
where $\mathbf{s^1}$ is a sentence in the $\mathcal{L}_1$ language and $\mathbf{s^2}$ its corresponding translation in $\mathcal{L}_2$. 

It is clear that the one provided is an extreme simplification of what a translator represents. However, approaching the problem in this way, it is intuitable that the estimation of the function f can be constructed using pre-trained language models. The details of such interpretation will be provided in later sections. 

\subsubsection{Resource Collection}
Building a suitable dataset for training a translation model is crucial for this task. The data to be collected must be texts in the low-resource language coupled, however, with corresponding translations in the high-resource language. One procedure for confirming the goodness of the collected translations is to have a domain expert view the dataset to ensure that no errors are present and that the material is of optimal quality for training a model. 

Currently, our project is conducting an extensive research phase to create a Sardinian-Italian dataset in which pieces of Sardinian text are paired with corresponding Italian translations. The choice of Italian as a high-resource language fell on the similarity in syntax and vocabulary between this language and Sardinian, and also, for obvious geographic-political reasons, such a translator would have a major impact on the enhancement of the language. 
This discussion is easily generalizable to any low-resource language. Hence, constructing a dataset that pairs texts in the source language with texts in the most similar high-resource language is recommended. 

\subsubsection{Resource Pre-Processing}
The data should be prepared for later use in the model training phase, as in the previous section. Since the collected texts will later be fed to a language model, it is necessary to use a tokenizer to transform the data into tokens in the most appropriate way for the model that will receive the input texts. There are several options for the tokenizer; on the one hand, a pre-trained tokenizer based on a language similar to the low-resource language can be adopted. On the other hand, the tokenizer can be directly trained for the target language \cite{atuhurra2024introducing}. 

As in previous cases, the dataset preprocessed with the described approaches will be split into two parts: one for the translation model training step and one for the evaluation step.
Currently, analyses are underway to determine whether, in the case of the Sardinian language, a tokenizer of the Italian language is sufficient to represent the entire vocabulary of the minority language or whether it is more convenient to train a dedicated tokenizer from scratch. The fact that there are no language models for Sardinian implies that there is no tokenizer for this language either, so it seems more appropriate to train it. 

\subsubsection{Model Development}
%
%Once the texts have been tokenized and the datasets have been prepared, we can move on to building a machine translation model. 
The following step is to define the machine translation model, which can be built through a fine-tuning phase of pre-trained LLMs \cite{zhang2023machine,zheng2024fine}; the approach aims to extend the knowledge of the LLMs through an additional training phase so that the neural network can recognize the patterns associated between the high resource language it already knows and the low resource language it does not know. With this strategy, the model can accurately translate from the intended language to the other (and vice versa). 

In the specific case of the Sardinian language, existing LLM-based bidirectional Sardinian-Italian and Italian-Sardinian machine translation models are currently being tested. At the moment, experiments are being carried out by exploiting minimal knowledge of the open-source language model Llama3; this model is fine-tuned in the translation task by receiving input texts written in both Sardinian and Italian. The choice of Italian as the reference language was made because this language turns out to be quite similar to the Sardinian language and remarkably familiar to Llama3; moreover, the choice was also guided by the available data since the largest number of available translations (needed to fine-tune the model) are in Sardinian and Italian.

\subsubsection{Model Evaluation}
Assess the quality of the translation produced by a model is extremely tricky to calculate; in fact, it may happen that although some words may be dissimilar to each other, the overall meaning of the sentence with which one is confronted remains unchanged, and the quality of the translation is high despite seemingly appearing to be the opposite. 
However, assessing context when dealing with poorly understood languages is challenging. Therefore, it is reasonable to identify a set of metrics that try to find a trade-off between form and substance. 
To this end, the training data may then be used to identify the benchmark metrics for the translation task. In our context, three main metrics have been selected. In particular, they allow us to understand how well the fine-tuned model has learned in translating the new language it did not know, whether given as input or requested as output. The identified metrics are 

\begin{itemize}
    \item \textbf{BiLingual Evaluation Understudy (BLEU)}: it measures the quality of a translation by comparing \textit{n-grams} with one or more reference translations, penalizing translations that are too short;
    \item \textbf{Translation Edit Rate (TER)}: it computes the minimum number of changes (insertions, deletions, substitutions) required to turn a machine translation into a reference translation, expressed as a percentage;
    \item \textbf{Metric for Evaluation of Translation with Explicit ORdering (METEOR)}: it evaluates translation quality based on not only n-gram correspondences but also synonyms, roots, and grammatical structure, giving greater weight to linguistic quality.
\end{itemize}

For the sake of completeness, \citet{lee2023survey} provide a deeper explanation of the mentioned metrics.

\subsection{Speech Processing Component}
The audio data processing module can be interpreted as a parallel module to the one described above; in fact, this module can almost be considered a framework component that can be detached without complicating the pipeline. Its role is to generate as much textual data as possible from the audio data; in short, it has a \textit{transcription} role. For the sake of completeness, the reverse process of transforming text chunks into spoken audio will also be described in this paper; although this tool does not add any value to the language model generation task, its use is essential for the preservation and enhancement of low-resource languages. 

\subsubsection{Problem Formalization}
The problem of turning audio into written text, also known as speech-to-text, can be formulated mathematically as a maximum likelihood problem. In detail, given an audio signal $A$, the goal is to find the most likely sequence of words $W$ that matches that signal. In formula:

\begin{equation}
\label{eq:1}
    W^* = \arg\max_W \mathbb{P}(A \mid W)\mathbb{P}(W)
\end{equation}

where $W^*$ is the optimal word sequence, $\mathbb{P}(A \mid W)$ is the probability that the audio signal $A$ is generated by the word sequence $W$, and $\mathbb{P}(W)$ the probability that the word sequence $W$ is grammatically and semantically correct in the language under consideration. Usually, the former probability is estimated by an acoustic model while the latter by a linguistic model.

Similarly, the problem of converting text to audio, commonly known as speech-to-text, can be formulated as: 

\begin{equation}
\label{eq:2}
    A^* = \arg\max_A \mathbb{P}(A \mid W)
\end{equation}

where $W$ represents a written text, $A$ the most likely audio for that word sequence, and $A^*$ the optimal audio signal. The value of $\mathbb{P}(A \mid W)$ is interpreted as the probability that the audio signal $A$ matches the text sequence $W$ and is most often computed from an acoustic synthesis model, trained to map the text to realistic and natural speech signals.

\subsubsection{Resource Collection}
As mentioned above, in order to estimate the probabilities in Formula \ref{eq:1} and Formula \ref{eq:2}, it is often necessary to use neural model training; thus, this requires careful and numerous audio data collection. In these cases, the datasets must contain audio signals and corresponding transcripts. 
Collecting such data is not straightforward for low-resource languages because audio signals often either do not possess transcripts or, if they do, are incorrect; a domain expert checking phase can help eliminate this criticality and make the collected material of high quality. 

However, it is possible on the Web to use public datasets that provide audio and transcripts even in minority languages, the \textit{Common Voice} project \cite{ardila2019common} being one example. The search for an audio-transcript dataset for the Sardinian language started with this source, which provides about 2 hours of Sardinian audio in multiple macro-variants, which is fully transcribed. 

\subsubsection{Resource Pre-Processing}
Whether the speech-to-text problem or the text-to-speech problem is being addressed, the collection phase must be accompanied by a nontrivial pre-processing phase in which, especially the audio files, are processed so that they can be used as model inputs. 
In the first case (s-t-t), the collected audio must be transformed into spectrograms in order to be used, as the algorithms that address this problem usually receive as input the spectrogram to estimate the probability described in Formula \ref{eq:1} and return a text-to-speech via a language model that estimates the second probability of the same Formula.

In the second case (t-t-s), the textual data must be tokenized, as described in the previous section, to be given as input to a model that estimates the probability of Formula \ref{eq:2}. This type of model returns as output an n-dimensional array that, through a synthesizer and selection of the correct frequency, is transformed into audio and can be audible by human ears. 

The data processed in the manner described in both cases must then be split into training and test sets to be used later for the model learning phase. The procedure described, with the Sardinian language datasets obtained from Common Voice, was also applied by our research group, and, as in the previous cases, the data were handled in Python through the \textit{datasets} library. 

\subsubsection{Model Development}
Although numerous models able to address the problems described above have been developed, existing pre-trained models cannot transcribe or generate audio in unfamiliar languages, such as low-resource languages. 
However, following the strategy used for the translation part, it is possible to fine-tune several LLMs that perform these tasks and teach them to perform them even for unfamiliar languages.
With the right setting, a large number of data, and the right number of training epochs, it is possible to fine-tune large models and make them capable of transcribing into new languages and synthesizing audio from text in these languages. 

In the context of our case study, some fine-tuning experiments were carried out with a pre-trained model of s-t-t, the Whisper model \cite{radford2022whisper}, which is characterized by knowledge of multiple languages and thus ease of learning new languages. The configuration used for the tuning is shown in Table \ref{tab:tuning_setting2}.

\begin{table}[h!] 
	\centering
	\caption{Setting used for fine-tuning the Whisper model for the s-t-t task. }
    \label{tab:tuning_setting2}
    \begin{tabular}{lc}
    \hline
        \textbf{Hyper-Parameter}& \textbf{Value}\\
    \hline
        Epochs & 15\\
        Batch size& 4\\
        Learning Rate&1e-5\\
        Weight decay&0.01\\
        Training Loss& Word Error Rate\\
    \hline
    \end{tabular}
\end{table}

Currently, the research has focused mainly on this type of task; a suitable model, among the existing ones, has not yet been identified for the t-t-s task; however, the choice of fine-tuning a pre-trained model remains consistent, even in this case. 

\subsubsection{Model Evaluation}
In the s-t-t task, the goal is to transcribe an audio signal into a text sequence correctly. In this context, the transcription quality can be assessed through various metrics that measure accuracy against a reference transcription. One of the most commonly used metrics for this task is the Word Error Rate (WER) \cite{park2008empirical}, a metric that measures the percentage of erroneous words in the transcript compared to the reference. In the case of the experiments we conducted for Sardinian, the smallest value of WER achieved was about 17. 

In the t-t-s task, the goal is to generate a natural, intelligible audio signal from a sequence of text, and the evaluation must focus on both objective and subjective aspects. The most common metric for subjective evaluation is the Mean Opinion Score \cite{viswanathan2005measuring}, calculated with human listeners rating the quality of the generated audio on an increasing scale from 1 to 5. For objective evaluation, one of the most representative metrics is Mel Cepstral Distortion (MCD) \cite{kominek2008synthesizer}, which measures the distance between the mel-cepstrum of the synthesized audio signal and that of the reference signal. 
Therefore, the previously constructed test datasets are used to calculate and optimize these metrics to ensure the goodness of fit of the constructed models. 

\subsection{Generative Modeling Component}
The final component of the framework proposed in this paper is the construction of an LLM capable of performing common generative AI tasks for a low-resource language. The tools built in the previous stages of the framework are aimed at generating as much high-quality data as possible to train the generative model. 

\subsubsection{Problem Formalization}
For the sake of clarity, the mathematical structure of a generative language model is as follows:
 
\begin{itemize}
    \item \textbf{Word Conditional Probability}: a generative model tries to approximate the conditional probability of a sequence of words $s_1,s_2,\dots,s_n$, where n is the length of the sequence. The total probability of a sequence is given by:
    \begin{equation}
        \mathbb{P}(s_1, s_2, \dots, s_n) = \prod_{j=1}^{n} \mathbb{P}(s_j \mid s_1, s_2, \dots, s_{j-1})
    \end{equation}
    where $\mathbb{P}(s_j \mid s_1, s_2, \dots, s_{j-1})$ represents the probability of the word $s_j$, given the previous sequence $s_1, s_2, \dots, s_{j-1}$. 
    \item \textbf{Neural Network}: a common way to model this conditional probability is to use a neural network. The model receives a partial sequence as input and estimates the probability distribution for the next word. 
In notation:

    \begin{equation}
        \begin{split}
        &h_j = f(h_{j-1}, s_{j-1};
        \textbf{\theta)} \\
        &\mathbb{P}(s_t \mid s_1, s_2, \dots, s_{j-1}) = \text{\textit{softmax}}(h_j)
        \end{split}
    \end{equation}

    where $h_j$ is the hidden state that contains the context information up to the word $s_{j-1}$, $f$ is a function (defined by the neural network) that updates the hidden state, \textbf{\theta)} are the parameters of the model and the \textit{softmax} function computes the probability for each vocabulary word.
    \item \textbf{Generation}: the text generation is done by sampling from the distribution
    \begin{equation}
        s_j \sim \mathbb{P}(s_j \mid s_1, s_2, \dots, s_{j-1})
    \end{equation}
    This process is repeated word by word until a complete text sequence is generated.
\end{itemize}

\subsubsection{Resource Collection}
The task of data collection for the creation of a generative model dedicated to a low-resource language has already been described throughout the paper. All or nearly all of the modules in the described framework were tasked with producing high-quality material to train a language model that knows best the main features of the language of interest. 

\subsubsection{Resource Pre-Processing}
Before training a generative language model, textual data must be preprocessed to be effectively used by the model. The first step is tokenization, a process described earlier, which involves breaking up the text into smaller units called tokens. The tokens can be words, subwords, or even characters, depending on the type of model being built. Next, the tokens must be converted into numbers through a vectorization process, often using a dictionary that assigns each token an integer number.

Another important aspect is managing rare or unknown words: these must be replaced with a special token or subword using proper techniques to break them down into smaller units. 
Finally, the length of input sequences must be managed. Sequences that are too long must be truncated, while those that are too short must be filled with padding tokens to ensure that all sequences are the same size, facilitating batch processing by the model.

\subsubsection{Model Development}
There are two main approaches to develop a generative language model for low-resource languages. The former, more common and simple, is fine-tuning a pre-trained model on a high-resource language. This approach takes advantage of the capability of pre-trained models, such as GPT or BERT, that have been trained on large amounts of multilingual or related language data. One starts with these pre-trained models and "fits" them to the low-resource language using available data produced with the tools described above. This process requires less data and computing power than training from scratch, allowing the model to generalize better.
The latter, a more complex, less common, but more effective, is to build a model from scratch. In this case, a neural network is created and trained directly on low-resource language data. This approach is feasible only if there is sufficient data; the effectiveness of from-scratch training over fine-tuning is only apparent in cases where large amounts of data are available to learn language structures well. This is why our framework was focused on generating new data for low-resource languages.

From our point of view, the best strategy to build a Sardinian language model is still being analyzed. The analysis is currently focusing on a quantitative assessment of the available data and the computational capacity available to train or tune the model. 

\subsubsection{Model Evaluation}
The evaluation of a generative language model trained on a new language can be approached with several strategies.
One of the most common methods is the use of automatic metrics, such as \textit{perplexity}, which measures how well the model predicts the next word in a sequence of text. Low perplexity indicates that the model generates text that is more consistent with real language. Other common metrics include the previously described BLEU, which compares generated text with reference texts.

Although the metrics described above are objective metrics that make it clear how well the model is performing, it is useful to supplement the analysis of the metrics with qualitative assessment through human reviewers who are experts in the target language. Annotators assess the \textit{fluency} and \textit{consistency} of the generated text concerning grammar and meaning in the language. In many cases, it is necessary to combine automatic metrics and human judgments to obtain a more reliable evaluation.

Finally, the generative model can also be evaluated on a variety of specific tasks, such as Named Entity Recognition (NER), text classification, or sentence completion, to test its ability to handle different tasks within the target language. This benchmarking approach provides insight into whether the model improves over previous solutions or benchmark models, offering a more comprehensive overview of its performance on different language aspects.

\section{Considerations and Recommendations}
\label{sec:considerations}
This section will provide a broader analysis of the framework developed for generating linguistic models for low-resource languages. We will examine the challenges and opportunities that arise from different perspectives: linguistic, technological, sustainability, and societal. By reflecting on the particular case of Sardinian, we aim to provide general recommendations that can be applied to other low-resource languages. These insights will guide future efforts in language preservation, technological innovation, and societal integration.

\subsection{Linguistic Perspective}
From a linguistic perspective, developing resources for resource-poor languages such as Sardinian presents several unique challenges. One of the main issues is the variability of dialects, which can hinder the creation of a unified model that accurately reflects the language as a whole. 
Creating a PoS tagger, translator, or speech-to-text system that can handle these variations requires a careful balance between \textit{generalization} and \textit{specificity}. Such a scenario is common to many low-resource languages, which often lack standardization in orthography and grammar.

However, the case of Sardinian also highlights the opportunity for a close relationship between minority languages and their cultural identity. By preserving and promoting Sardinian through linguistic technologies, we also aim to preserve the unique cultural heritage embedded in the language. It is essential to involve native speakers in the development process to ensure linguistic accuracy and to capture the richness of the language in its authentic context. For other low-resource languages, linguistic variation and cultural significance must be taken into account in the design of language technologies to ensure that these systems remain relevant and useful to the communities they are intended to serve.

\subsection{Technological Perspective}
The technological perspective on developing linguistic models for low-resource languages, such as Sardinian, highlights several critical factors. First, data availability is a major challenge; low-resource languages often lack substantial digital corpora, which are essential for training ML models. This scarcity requires innovative data collection methods, including crowd-sourcing, partnerships with educational institutions, and leveraging existing resources such as literature and oral traditions. With its rich oral history, the Sardinian language presents an opportunity to create unique datasets encompassing various dialects, ensuring that the resulting technologies reflect the language's true linguistic diversity.

Moreover, the choice of technology is paramount. Traditional NLP tools may not be directly applicable to low-resource languages due to their specific linguistic features. Adopting adaptable, modular architectures in the development process can significantly enhance the effectiveness of language technologies for Sardinian and similar languages. For example, using transfer learning techniques allows for the incorporation of models pre-trained on high-resource languages, which can be fine-tuned with the limited Sardinian data available. This approach not only maximizes resource efficiency but also ensures the robustness of the resulting applications. Overall, leveraging modern technological advancements while considering the unique characteristics of low-resource languages is crucial for developing effective linguistic models.

\subsection{Sustainability Perspective}
From a sustainability perspective, developing linguistic models for low-resource languages carries significant implications for preserving and promoting linguistic diversity. Sustainability in this context refers not only to the environmental impact of technological developments but also to the cultural and linguistic vitality of minority languages. For Sardinian, integrating technology in language preservation efforts can foster a sustainable ecosystem where the language continues to thrive alongside modern advancements.

A key aspect of sustainability is the need for ongoing community engagement and capacity building. By involving local speakers in the development of language technologies, a sense of ownership is created, and it is ensured that the tools produced are genuinely relevant to their needs. Furthermore, training native speakers in digital literacy and technology use can empower them to contribute actively to the ongoing development and maintenance of linguistic resources. This approach helps to mitigate the risk of technological solutions becoming disconnected from the communities they aim to serve.

In addition, fostering a sustainable relationship between technology and Sardinian means promoting long-term support for language preservation initiatives. This includes securing funding for ongoing projects, creating partnerships with educational institutions, and establishing structures for collaboration among stakeholders. The relevance and effectiveness of technologies developed for low-resource languages can be maintained by ensuring that resources are continuously updated and improved. A sustainable approach to language technology not only preserves languages such as Sardinian but also increases their resilience in a rapidly changing world.

\subsection{Society Perspective}
The social perspective on the development of LLMs for low-resource languages emphasizes the crucial role of language in cultural identity and community cohesion. Sardinian, for example, as well as other minority languages, is not just a means of communication; it embodies the history, traditions, and values of its speakers. Therefore, fostering its use through technological interventions can significantly improve social ties and promote cultural pride among younger generations.

In addition, the accessibility of language technologies, such as PoS tagging, translation, and speech processing systems, can strengthen communities by providing education, communication, and cultural exchange tools. By integrating these technologies into daily life, we can encourage the use of Sardinian in various areas, including education, media, and public administration. This integration can help combat the perception of minority languages as obsolete or irrelevant, reinforcing their significance in a modern context.

Community involvement is crucial in this regard; initiatives involving local speakers in creating and using these technologies can foster a sense of belonging and encourage active participation in cultural preservation. In addition, promoting awareness of the importance of linguistic diversity can encourage broader social support for language revitalization efforts. Thus, a social approach to language technology not only supports the Sardinian language and all low-resource languages but also contributes to the broader goals of social equity and cultural sustainability.

\section{Conlusions and Future Work}
\label{sec:conclusions}
In conclusion, the framework described for generating language models for low-resource languages presents a multifaceted approach to addressing the unique challenges of languages such as Sardinian. Through careful consideration of linguistic, technological, sustainability, and social perspectives, we have outlined strategies that can improve language preservation and promote cultural identity.

Future work will focus on developing a more detailed and consistent framework than the one presented in this article and on new ways to expand the dataset for Sardinian and other low-resource languages. In addition, continued collaboration with local communities will be essential to ensure that the tools being developed are relevant and valuable. Exploring partnerships with educational institutions can provide avenues for integrating language technologies into the curriculum, fostering new generations of speakers and advocates.
In addition, lessons learned from the Sardinian case study can serve as a model for other low-resource languages, encouraging a comprehensive approach to language preservation that values diversity and promotes inclusiveness. By continuing to innovate and invest in these efforts, it is possible to work toward a future in which low-resource languages are not only preserved but also celebrated as vital components of our shared human heritage.

\newpage
\bibliographystyle{plainnat}
\bibliography{references}

\begin{thebibliography}{63}
\providecommand{\natexlab}[1]{#1}
\providecommand{\url}[1]{\texttt{#1}}
\expandafter\ifx\csname urlstyle\endcsname\relax
  \providecommand{\doi}[1]{doi: #1}\else
  \providecommand{\doi}{doi: \begingroup \urlstyle{rm}\Url}\fi

\bibitem[Ardila et~al.(2019)Ardila, Branson, Davis, Henretty, Kohler, Meyer, Morais, Saunders, Tyers, and Weber]{ardila2019common}
Rosana Ardila, Megan Branson, Kelly Davis, Michael Henretty, Michael Kohler, Josh Meyer, Reuben Morais, Lindsay Saunders, Francis~M Tyers, and Gregor Weber.
\newblock Common voice: A massively-multilingual speech corpus.
\newblock \emph{arXiv preprint arXiv:1912.06670}, 2019.

\bibitem[Atuhurra et~al.(2024)Atuhurra, Shindo, Kamigaito, and Watanabe]{atuhurra2024introducing}
Jesse Atuhurra, Hiroyuki Shindo, Hidetaka Kamigaito, and Taro Watanabe.
\newblock Introducing syllable tokenization for low-resource languages: A case study with swahili.
\newblock \emph{arXiv preprint arXiv:2406.15358}, 2024.

\bibitem[Bansal et~al.(2018)Bansal, Kamper, Livescu, Lopez, and Goldwater]{bansal2018low}
Sameer Bansal, Herman Kamper, Karen Livescu, Adam Lopez, and Sharon Goldwater.
\newblock Low-resource speech-to-text translation.
\newblock \emph{arXiv preprint arXiv:1803.09164}, 2018.

\bibitem[Baziotis et~al.(2020)Baziotis, Haddow, and Birch]{baziotis2020language}
Christos Baziotis, Barry Haddow, and Alexandra Birch.
\newblock Language model prior for low-resource neural machine translation.
\newblock \emph{arXiv preprint arXiv:2004.14928}, 2020.

\bibitem[Bird(2020)]{bird2020decolonising}
Steven Bird.
\newblock Decolonising speech and language technology.
\newblock In \emph{28th International Conference on Computational Linguistics, COLING 2020}, pages 3504--3519. Association for Computational Linguistics (ACL), 2020.

\bibitem[Buys and Botha(2016)]{buys2016cross}
Jan Buys and Jan~A Botha.
\newblock Cross-lingual morphological tagging for low-resource languages.
\newblock \emph{arXiv preprint arXiv:1606.04279}, 2016.

\bibitem[Cahyawijaya et~al.(2024)Cahyawijaya, Lovenia, and Fung]{cahyawijaya2024llms}
Samuel Cahyawijaya, Holy Lovenia, and Pascale Fung.
\newblock Llms are few-shot in-context low-resource language learners.
\newblock \emph{arXiv preprint arXiv:2403.16512}, 2024.

\bibitem[Cardenas et~al.(2019)Cardenas, Lin, Ji, and May]{cardenas2019grounded}
Ronald Cardenas, Ying Lin, Heng Ji, and Jonathan May.
\newblock A grounded unsupervised universal part-of-speech tagger for low-resource languages.
\newblock \emph{arXiv preprint arXiv:1904.05426}, 2019.

\bibitem[Christodoulopoulos et~al.(2010)Christodoulopoulos, Goldwater, and Steedman]{christodoulopoulos2010two}
Christos Christodoulopoulos, Sharon Goldwater, and Mark Steedman.
\newblock Two decades of unsupervised pos induction: How far have we come?
\newblock In \emph{Proceedings of the 2010 Conference on Empirical Methods in Natural Language Processing}, pages 575--584, 2010.

\bibitem[Cieri et~al.(2016)Cieri, Maxwell, Strassel, and Tracey]{cieri2016selection}
Christopher Cieri, Mike Maxwell, Stephanie Strassel, and Jennifer Tracey.
\newblock Selection criteria for low resource language programs.
\newblock In \emph{Proceedings of the Tenth International Conference on Language Resources and Evaluation (LREC'16)}, pages 4543--4549, 2016.

\bibitem[de~Jesus and Nunes(2024)]{de2024data}
Gabriel de~Jesus and S{\'e}rgio~Sobral Nunes.
\newblock Data collection pipeline for low-resource languages: A case study on constructing a tetun text corpus.
\newblock In \emph{Proceedings of the 2024 Joint International Conference on Computational Linguistics, Language Resources and Evaluation (LREC-COLING 2024)}, pages 4368--4380, 2024.

\bibitem[Devlin(2018)]{devlin2018bert}
Jacob Devlin.
\newblock Bert: Pre-training of deep bidirectional transformers for language understanding.
\newblock \emph{arXiv preprint arXiv:1810.04805}, 2018.

\bibitem[Dunning(1994)]{dunning1994statistical}
Ted Dunning.
\newblock \emph{Statistical identification of language}.
\newblock Computing Research Laboratory, New Mexico State University Las Cruces, 1994.

\bibitem[Duong et~al.(2014)Duong, Cohn, Verspoor, Bird, and Cook]{duong2014can}
Long Duong, Trevor Cohn, Karin Verspoor, Steven Bird, and Paul Cook.
\newblock What can we get from 1000 tokens? a case study of multilingual pos tagging for resource-poor languages.
\newblock In \emph{Proceedings of the 2014 Conference on Empirical Methods in Natural Language Processing (EMNLP)}, pages 886--897, 2014.

\bibitem[Fang and Cohn(2016)]{fang2016learning}
Meng Fang and Trevor Cohn.
\newblock Learning when to trust distant supervision: An application to low-resource pos tagging using cross-lingual projection.
\newblock \emph{arXiv preprint arXiv:1607.01133}, 2016.

\bibitem[Fang and Cohn(2017)]{fang2017model}
Meng Fang and Trevor Cohn.
\newblock Model transfer for tagging low-resource languages using a bilingual dictionary.
\newblock \emph{arXiv preprint arXiv:1705.00424}, 2017.

\bibitem[Ferrer et~al.(2017)Ferrer, Koch, and Marzo]{marzo17}
Eduardo~Blasco Ferrer, Peter Koch, and Daniela Marzo, editors.
\newblock \emph{Manuale di linguistica sarda}.
\newblock De Gruyter, Berlin, Boston, 2017.
\newblock ISBN 9783110274615.
\newblock \doi{doi:10.1515/9783110274615}.
\newblock URL \url{https://doi.org/10.1515/9783110274615}.

\bibitem[Ghimenton and Depau(2016)]{ghimenton2016ideologies}
Anna Ghimenton and Giovanni Depau.
\newblock Ideologies and expressed attitudes in internet: Comparing ethnic identities in two regional communities (veneto and sardinia).
\newblock \emph{Vanishing Languages in Context. Ideological, Attitudinal and Social Identity Perspectives}, 114:\penalty0 73--102, 2016.

\bibitem[Grandini et~al.(2020)Grandini, Bagli, and Visani]{grandini2020metrics}
Margherita Grandini, Enrico Bagli, and Giorgio Visani.
\newblock Metrics for multi-class classification: an overview.
\newblock \emph{arXiv preprint arXiv:2008.05756}, 2020.

\bibitem[Kazantsevaa et~al.(2024)Kazantsevaa, Kuhna, Larkina, Littella, Lothiana, Akwirat{\'e}kha’Martina, Tessiera, Valentini-Botinhaoc, Wellsc, and Yamagishib]{kazantsevaaspeech}
Ross~Krekoskid Kazantsevaa, Roland Kuhna, Samuel Larkina, Patrick Littella, Delaney Lothiana, Korin~Richmondc Akwirat{\'e}kha’Martina, Marc Tessiera, Cassia Valentini-Botinhaoc, Dan Wellsc, and Junichi Yamagishib.
\newblock Speech generation for indigenous language education.
\newblock \emph{Computer Speech \& Language}, 2024.

\bibitem[Khanna et~al.(2021)Khanna, Washington, Tyers, Bayatl{\i}, Swanson, Pirinen, Tang, and Alos~i Font]{khanna2021recent}
Tanmai Khanna, Jonathan~N Washington, Francis~M Tyers, Sevilay Bayatl{\i}, Daniel~G Swanson, Tommi~A Pirinen, Irene Tang, and Hector Alos~i Font.
\newblock Recent advances in apertium, a free/open-source rule-based machine translation platform for low-resource languages.
\newblock \emph{Machine Translation}, 35\penalty0 (4):\penalty0 475--502, 2021.

\bibitem[King(2015)]{king2015practical}
Benjamin~Philip King.
\newblock \emph{Practical Natural Language Processing for Low-Resource Languages.}
\newblock PhD thesis, University of Michigan, 2015.

\bibitem[Kominek et~al.(2008)Kominek, Schultz, and Black]{kominek2008synthesizer}
John Kominek, Tanja Schultz, and Alan~W Black.
\newblock Synthesizer voice quality of new languages calibrated with mean mel cepstral distortion.
\newblock In \emph{SLTU}, pages 63--68, 2008.

\bibitem[Kumar et~al.(2019)Kumar, Jha, and Sahula]{kumar2019augmented}
Rashi Kumar, Piyush Jha, and Vineet Sahula.
\newblock An augmented translation technique for low resource language pair: Sanskrit to hindi translation.
\newblock In \emph{Proceedings of the 2019 2nd international conference on algorithms, computing and artificial intelligence}, pages 377--383, 2019.

\bibitem[Lankford et~al.(2023)Lankford, Afli, and Way]{lankford2023adaptmllm}
S{\'e}amus Lankford, Haithem Afli, and Andy Way.
\newblock adaptmllm: Fine-tuning multilingual language models on low-resource languages with integrated llm playgrounds.
\newblock \emph{Information}, 14\penalty0 (12):\penalty0 638, 2023.

\bibitem[Lee et~al.(2023)Lee, Lee, Moon, Park, Seo, Eo, Koo, and Lim]{lee2023survey}
Seungjun Lee, Jungseob Lee, Hyeonseok Moon, Chanjun Park, Jaehyung Seo, Sugyeong Eo, Seonmin Koo, and Heuiseok Lim.
\newblock A survey on evaluation metrics for machine translation.
\newblock \emph{Mathematics}, 11\penalty0 (4):\penalty0 1006, 2023.

\bibitem[Linzmeier et~al.(2021)Linzmeier, Pisano, et~al.]{linzmeier2021visibilita}
Laura Linzmeier, Simone Pisano, et~al.
\newblock Visibilit{\`a} delle variet{\`a} italo-romanze nel paesaggio linguistico della sardegna settentrionale e nel cyberspazio: il caso del sassarese e del gallurese.
\newblock \emph{La presenza dei dialetti italo-romanzi nel paesaggio linguistico. Ricerche e Riflessioni}, pages 109--130, 2021.

\bibitem[Madaan et~al.(2020)Madaan, Rijhwani, Anastasopoulos, Yang, and Neubig]{madaan2020practical}
Aman Madaan, Shruti Rijhwani, Antonios Anastasopoulos, Yiming Yang, and Graham Neubig.
\newblock Practical comparable data collection for low-resource languages via images.
\newblock \emph{arXiv preprint arXiv:2004.11954}, 2020.

\bibitem[Magueresse et~al.(2020)Magueresse, Carles, and Heetderks]{magueresse2020low}
Alexandre Magueresse, Vincent Carles, and Evan Heetderks.
\newblock Low-resource languages: A review of past work and future challenges.
\newblock \emph{arXiv preprint arXiv:2006.07264}, 2020.

\bibitem[Makarov and Clematide(2018)]{makarov2018neural}
Peter Makarov and Simon Clematide.
\newblock Neural transition-based string transduction for limited-resource setting in morphology.
\newblock In \emph{Proceedings of the 27th International Conference on Computational Linguistics}, pages 83--93, 2018.

\bibitem[Marra(2012)]{marra2012lingue}
Maria~Antonietta Marra.
\newblock Lingue di minoranza a scuola: uno sguardoalla sardegna a dieci anni dalla legge 482/99.
\newblock In \emph{Linguistica educativa: atti del XLIV Congresso internazionale di studi della Societ{\`a} di linguistica italiana (SLI): Viterbo, 27-29 settembre 2010.-(Pubblicazioni della Societ{\`a} linguistica italiana; 55)}, pages 249--266. Bulzoni, 2012.

\bibitem[Masullo et~al.(2021)Masullo, Castelli, Meloni, Meluzzi, et~al.]{masullo2021dialetti}
Camilla Masullo, Claudia Castelli, Cinzia Meloni, Chiara Meluzzi, et~al.
\newblock Dialetti su instagram: usi, differenze e atteggiamenti linguistici.
\newblock \emph{BIBLIOTECA DI LINGUISTICA E FILOLOGIA}, 7:\penalty0 237--254, 2021.

\bibitem[Mensching(2000)]{mensching2000internet}
Guido Mensching.
\newblock The internet as a rescue tool of endangered languages: Sardinian.
\newblock In \emph{Proceeding Conference Multilinguae: multimedia and minority languages. San Sebastian: The Association of Electronics and Information Technology Industries}, 2000.

\bibitem[Mercer et~al.(1992)]{mercer1992class}
Robert~L Mercer et~al.
\newblock Class-based n-gram models of natural language.
\newblock \emph{Computational Linguistics}, 18\penalty0 (4):\penalty0 18--4, 1992.

\bibitem[Mereu(2021)]{mereu2021efforts}
Daniela Mereu.
\newblock Efforts to standardise minority languages. the case of sardinian.
\newblock \emph{Europ{\"a}isches Journal f{\"u}r Minderheitenfragen}, 14\penalty0 (1-2):\penalty0 76--95, 2021.

\bibitem[Merlo and normale~superiore (Italy)(2009)]{merlo2009italia}
C.~Merlo and Scuola normale~superiore (Italy).
\newblock \emph{L'Italia dialettale: rivista di dialettologia italiana}.
\newblock L'Italia dialettale. Arti Grafiche Pacini Mariotti, 2009.
\newblock URL \url{https://books.google.it/books?id=3h_DAr8XOOEC}.

\bibitem[Mohanty et~al.(2024)Mohanty, Dash, and Parida]{mohanty2024applying}
Sushree~Sangita Mohanty, Satya~Ranjan Dash, and Shantipriya Parida.
\newblock \emph{Applying AI-based Tools and Technologies Towards Revitalization of Indigenous and Endangered Languages}.
\newblock Springer, 2024.

\bibitem[Mura(2024)]{mura2024attitudes}
Piergiorgio Mura.
\newblock Attitudes towards sardinian and italian finally compared via the matched-guise technique.
\newblock \emph{International Journal of the Sociology of Language}, 2024\penalty0 (288):\penalty0 121--147, 2024.

\bibitem[Mura et~al.(July 3-6, 2024)Mura, Pisano, Carta, Giuliani, and Manca]{mura2024}
Piergiorgio Mura, Simone Pisano, Salvatore Carta, Alessandro Giuliani, and Manolo Manca.
\newblock \emph{The corpus of Sardinian emigrants:a tool for a quantitative approach to contact phenomena}.
\newblock MiLES: Minority Languages in European Societies - International Conference-Turin / Bard - BOOK OF ABSTRACTS. July 3-6, 2024.

\bibitem[Muzoora et~al.(2014)Muzoora, Terry, and Asiimwe]{muzoora2014valorisation}
Michael Muzoora, Daniel~R Terry, and Agatha~A Asiimwe.
\newblock The valorisation of african languages and policies in the african education systems: A case of uganda.
\newblock \emph{Universal Journal of Educational Research}, 2\penalty0 (1):\penalty0 42--50, 2014.

\bibitem[Olko and Sallabank(2021)]{olko2021revitalizing}
Justyna Olko and Julia Sallabank.
\newblock \emph{Revitalizing endangered languages: A practical guide}.
\newblock Cambridge University Press, 2021.

\bibitem[Parida et~al.(2024)Parida, Panwar, Lata, Mishra, and Sekhar]{parida2024building}
Shantipriya Parida, Shakshi Panwar, Kusum Lata, Sanskruti Mishra, and Sambit Sekhar.
\newblock Building pre-train llm dataset for the indic languages: a case study on hindi.
\newblock \emph{arXiv preprint arXiv:2407.09855}, 2024.

\bibitem[Park et~al.(2008)Park, Patwardhan, Visweswariah, and Gates]{park2008empirical}
Youngja Park, Siddharth Patwardhan, Karthik Visweswariah, and Stephen~C Gates.
\newblock An empirical analysis of word error rate and keyword error rate.
\newblock In \emph{Interspeech}, volume 2008, pages 2070--2073, 2008.

\bibitem[Paul et~al.(2024)Paul, Buckchash, Parida, and Prasad]{paul2024towards}
Ronny Paul, Himanshu Buckchash, Shantipriya Parida, and Dilip~K Prasad.
\newblock Towards a more inclusive ai: Progress and perspectives in large language model training for the s$\backslash$'ami language.
\newblock \emph{arXiv preprint arXiv:2405.05777}, 2024.

\bibitem[Pellegrini(1977)]{pellegrini1977carta}
Giovan~Battista Pellegrini.
\newblock Carta dei dialetti d'italia.
\newblock \emph{(No Title)}, 1977.

\bibitem[Pisano et~al.(2022)Pisano, Piunno, Ganfi, et~al.]{pisano2022appunti}
Simone Pisano, Valentina Piunno, Vittorio Ganfi, et~al.
\newblock Appunti per un corpus di sardo multimediale.
\newblock In \emph{Per una pianificazione del plurilinguismo in Sardegna}, pages 147--164. Condaghes, 2022.

\bibitem[Poutsma(2002)]{poutsma2002applying}
Arjen Poutsma.
\newblock Applying monte carlo techniques to language identification.
\newblock In \emph{Computational Linguistics in the Netherlands 2001}, pages 179--189. Brill, 2002.

\bibitem[Radford et~al.(2022)Radford, Kim, Xu, Brockman, McLeavey, and Sutskever]{radford2022whisper}
Alec Radford, Jong~Wook Kim, Tao Xu, Greg Brockman, Christine McLeavey, and Ilya Sutskever.
\newblock Robust speech recognition via large-scale weak supervision, 2022.
\newblock URL \url{https://arxiv.org/abs/2212.04356}.

\bibitem[Russo et~al.(2016)Russo, Pisano, and Soria]{russo2016sardinian}
Irene Russo, Simone Pisano, and Claudia Soria.
\newblock Sardinian on facebook: Analysing diatopic varieties through translated lexical lists.
\newblock In \emph{CLiC-it/EVALITA}, 2016.

\bibitem[Stoian et~al.(2020)Stoian, Bansal, and Goldwater]{stoian2020analyzing}
Mihaela~C Stoian, Sameer Bansal, and Sharon Goldwater.
\newblock Analyzing asr pretraining for low-resource speech-to-text translation.
\newblock In \emph{ICASSP 2020-2020 IEEE International Conference on Acoustics, Speech and Signal Processing (ICASSP)}, pages 7909--7913. IEEE, 2020.

\bibitem[Takeoka et~al.(2021)Takeoka, Akimoto, and Oyamada]{takeoka2021low}
Kunihiro Takeoka, Kosuke Akimoto, and Masafumi Oyamada.
\newblock Low-resource taxonomy enrichment with pretrained language models.
\newblock In \emph{Proceedings of the 2021 Conference on Empirical Methods in Natural Language Processing}, pages 2747--2758, 2021.

\bibitem[Tu et~al.(2019)Tu, Chen, Yeh, and Lee]{tu2019end}
Tao Tu, Yuan-Jui Chen, Cheng-chieh Yeh, and Hung-Yi Lee.
\newblock End-to-end text-to-speech for low-resource languages by cross-lingual transfer learning.
\newblock \emph{arXiv preprint arXiv:1904.06508}, 2019.

\bibitem[Tyers et~al.(2017)Tyers, i~Font, Fronteddu, and Mart{\'\i}n-Mor]{tyers2017rule}
Francis~M Tyers, H{\`e}ctor~Al{\`o}s i~Font, Gianfranco Fronteddu, and Adri{\`a} Mart{\'\i}n-Mor.
\newblock Rule-based machine translation for the italian-sardinian language pair.
\newblock \emph{The Prague Bulletin of Mathematical Linguistics}, 108\penalty0 (1):\penalty0 221, 2017.

\bibitem[Valdes(2007)]{valdes2007capitolo}
M~Valdes.
\newblock Capitolo secondo: Valori, opinioni e atteggiamenti verso le lingue locali.
\newblock \emph{Le lingue dei sardi-Una ricerca sociolinguistica}, pages 46--64, 2007.

\bibitem[Virdis(2003)]{virdis03a}
Maurizio Virdis.
\newblock La lingua sarda fra le lingue neolatine. storia uso e problemi.
\newblock In \emph{Convegno “La lingua e la cultura della Sardegna”}, 2003.
\newblock URL \url{https://www.academia.edu/22772692/La_lingua_sarda_fra_le_lingue_neolatine_Storia_uso_e_problemi}.

\bibitem[Viswanathan and Viswanathan(2005)]{viswanathan2005measuring}
Mahesh Viswanathan and Madhubalan Viswanathan.
\newblock Measuring speech quality for text-to-speech systems: development and assessment of a modified mean opinion score (mos) scale.
\newblock \emph{Computer speech \& language}, 19\penalty0 (1):\penalty0 55--83, 2005.

\bibitem[Wagner and Virchowstiftung(1921)]{wagner1921ländliche}
M.L. Wagner and Rudolf Virchowstiftung.
\newblock \emph{Das l{\"a}ndliche Leben Sardiniens im Spiegel der Sprache: kulturhistorisch-sprachliche Untersuchungen}.
\newblock W{\"o}rter und Sachen : Kulturhistorische Zeitschrift f{\"u}r Sprach- und Sachforschung. Beiheft 4. na, 1921.
\newblock URL \url{https://books.google.it/books?id=K6FIAQAAMAAJ}.

\bibitem[Xie et~al.(2023)Xie, Santurkar, Ma, and Liang]{xie2023data}
Sang~Michael Xie, Shibani Santurkar, Tengyu Ma, and Percy~S Liang.
\newblock Data selection for language models via importance resampling.
\newblock \emph{Advances in Neural Information Processing Systems}, 36:\penalty0 34201--34227, 2023.

\bibitem[Yarowsky et~al.(2001)Yarowsky, Ngai, and Wicentowski]{yarowsky-etal-2001-inducing}
David Yarowsky, Grace Ngai, and Richard Wicentowski.
\newblock Inducing multilingual text analysis tools via robust projection across aligned corpora.
\newblock In \emph{Proceedings of the First International Conference on Human Language Technology Research}, 2001.
\newblock URL \url{https://aclanthology.org/H01-1035}.

\bibitem[Yusupujiang and Ginzburg(2021)]{yusupujiang2021data}
Zulipiye Yusupujiang and Jonathan Ginzburg.
\newblock Data collection design for dialogue systems for low-resource languages.
\newblock \emph{Conversational Dialogue Systems for the Next Decade}, pages 387--392, 2021.

\bibitem[Zhang et~al.(2023)Zhang, Rajabi, Duh, and Koehn]{zhang2023machine}
Xuan Zhang, Navid Rajabi, Kevin Duh, and Philipp Koehn.
\newblock Machine translation with large language models: Prompting, few-shot learning, and fine-tuning with qlora.
\newblock In \emph{Proceedings of the Eighth Conference on Machine Translation}, pages 468--481, 2023.

\bibitem[Zheng et~al.(2024)Zheng, Hong, Wang, Su, Liang, and Wu]{zheng2024fine}
Jiawei Zheng, Hanghai Hong, Xiaoli Wang, Jingsong Su, Yonggui Liang, and Shikai Wu.
\newblock Fine-tuning large language models for domain-specific machine translation.
\newblock \emph{arXiv preprint arXiv:2402.15061}, 2024.

\bibitem[Zoph et~al.(2016)Zoph, Yuret, May, and Knight]{zoph2016transfer}
Barret Zoph, Deniz Yuret, Jonathan May, and Kevin Knight.
\newblock Transfer learning for low-resource neural machine translation.
\newblock \emph{arXiv preprint arXiv:1604.02201}, 2016.

\end{thebibliography}

\end{document}